\documentclass[sigconf]{acmart} 

\AtBeginDocument{%
  }

\setcopyright{acmlicensed}
\copyrightyear{2025} 
\acmYear{2025} 
\setcopyright{acmlicensed}
\acmConference[Websci '25]{Proceedings of the 17th ACM Web Science Conference 2025}{May 20--24, 2025}{New Brunswick, NJ, USA} 
\acmBooktitle{Proceedings of the 17th ACM Web Science Conference 2025 (Websci '25), May 20--24, 2025, New Brunswick, NJ, USA} 
\acmDOI{10.1145/3717867.3717896} \acmISBN{979-8-4007-1483-2/2025/05}

\begin{document}
\title{GenAI vs. Human Fact-Checkers: Accurate Ratings, Flawed Rationales}

\author{Yuehong Cassandra Tai}
\email{yhcasstai@psu.edu}
\orcid{1234-5678-9012}

\affiliation{%
  \institution{Penn State University}
  \city{State College}
  \state{PA}
  \country{USA}
}

\author{Khushi Navin Patni}
\affiliation{%
  \institution{Penn State University}
  \city{State College}
  \country{PA}}
\email{USA}

\author{Nicholas Daniel Hemauer}
\affiliation{%
  \institution{Penn State University}
  \city{State College}
  \country{PA}}
\email{USA}

\author{Bruce Desmarais}
\affiliation{%
  \institution{Penn State University}
  \city{State College}
  \state{PA}
  \country{USA}}

  \author{Yu-Ru Lin}
\affiliation{%
  \institution{University of Pittsburgh}
  \city{Pittsburgh}
  \state{PA}
  \country{USA}}

\renewcommand{\shortauthors}{Tai et al.}

\begin{abstract}
Despite recent advances in understanding the capabilities and limits of generative artificial intelligence (GenAI) models, we are just beginning to understand their capacity to assess and reason about the veracity of content. We evaluate multiple GenAI models across tasks that involve the rating of, and reasoning about, the credibility of information. The information in our experiments comes from content that subnational U.S. politicians post to Facebook. We find that GPT-4o, one of the most used AI models in consumer applications, outperforms other models, but all models exhibit only moderate agreement with human coders. Importantly, even when GenAI models accurately identify low-credibility content, their reasoning relies heavily on linguistic features and ``hard'' criteria, such as the level of detail, source reliability and language formality, rather than an understanding of veracity. We also assess the effectiveness of summarized versus full content inputs, finding that summarized content holds promise for improving efficiency without sacrificing accuracy. While GenAI has the potential to support human fact-checkers in scaling misinformation detection, our results caution against relying solely on these models.
\end{abstract}
\vspace{-5mm}

\ccsdesc[300]{Human-centered computing}
\ccsdesc[500]{Social and professional topics}
\vspace{-5mm}
\keywords{GenAI, fact-checker, LLM, misinformation, Facebook}

\maketitle

\section{Introduction}

Generative Artificial Intelligence (GenAI) has transformed workflows across industries and academia since its surge in popularity in 2022 \cite{abels2023chatgpt}. In scientific research, its adoption has accelerated at an unprecedented pace \cite{zhao2023survey}, demonstrating transformative potential in diverse applications. Researchers have successfully employed GenAI models to impersonate survey respondents \cite{bisbee2024synthetic}, generate counterfactual images for experimental research \cite{davidson2024start}, annotate text with human-comparable accuracy\cite{gilardi2023chatgpt}, identify conspiracy theories \cite{diab2024classifying}, and preemptively debunk election misinformation \cite{linegar2024prebunking}. Among these diverse applications of GenAI, a fundamental question remains about its ability to assess content credibility.

We assess the capacity of GenAI models to rate the veracity of content. The online proliferation of unreliable and misleading information, including misinformation, poses threats to democracies and societies, spurring political violence, undermining trust in democratic institutions globally, and endangering public health \cite{eliassi2020science, lazer2018science}. Combating misinformation requires reliable detection tools; however, the complexity of fact-checking remains largely beyond the capabilities of even state-of-the-art AI systems \cite{neumann2024diverse}. Given these limitations, one well-established alternative is to assess content reliability by using the credibility of source domains as a proxy \cite{lasser2023alternative, guess2020exposure}. This domain-based approach raises concerns about false positives, as not all content from unreliable domains is misinformation. 

Scholars have explored the potential of large language models (LLMs) as fact-checkers, building on foundational work showing that language models can verify claims without external knowledge bases \cite{lee2020language}. For example, using datasets like news headlines labeled for credibility \cite{gabriel2021misinfo}, researchers found that GenAI models without fine-tuning achieve moderate performance in detecting political misinformation \cite{ziems2024can}. This suggests that if GenAI can reliably assess credibility through zero-shot prompting, it could alleviate human resource demands in large-scale fact-checking. To advance this line of inquiry, we explore the perceived reasoning patterns that emerge during content credibility assessment in zero-shot settings, offering insights into the internal processes of GenAI. Specifically, we address the following research questions:

\noindent {\bf RQ1 (Viability):} Can GenAI models match or exceed the reliability of human coders? \\
\noindent {\bf RQ2 (Efficiency):} Can summaries generated by zero-shot GenAI models provide efficient credibility ratings? \\
\noindent {\bf RQ3 (Functionality):} How do GenAI models reason about information credibility?

To answer these questions, we analyzed text and video content from low-credibility domain sources that were shared on Facebook by state legislators in the U.S. This is a highly salient application, as it draws upon real-world content that has been disseminated by public figures. After achieving high intercoder reliability among human coders, we conducted an experiment to evaluate the credibility rating and reasoning capacity of multiple GenAI models via zero-shot prompting, including GPT 4o, Llama 3.1-8b and 70b, Gemma2-9b, and Flan-T5-XL. Gemma2-9b demonstrates comparable performance to GPT-4o in rating content, and Gemma2-9b and Llama 3.1-70b outperform other models when using summaries. Agreement with human coders is generally low to moderate across all models, suggesting that GenAI cannot be solely relied upon for fact-checking. Our reasoning analysis shows that models primarily rely on ``hard''criteria, such as the level of detail, source and reference quality, and linguistic features like formality and objectivity, rather than a true understanding of veracity. This reliance raises concerns about their ability to detect low-credibility content when sensational language is avoided, and selective ``details'' are strategically included. Therefore, we recommend hybrid human-AI approaches, especially in reasoning tasks.
\vspace{-.2cm}
\section{Method}

Despite the ongoing debate surrounding the definition of misinformation, we adopt a broad conceptualization, encompassing both fake content and misleading framing of factual content. This approach aligns with scholarly consensus that both forms constitute misinformation \cite{science_misinformation}. For example, a video titled ``Fauci Admits On Camera That He Lied About Masks'' with the blurb ``Wow, this is getting crazy'' was labeled as low-credibility because it misrepresented Dr. Fauci’s statements on mask use\footnote{https://rumble.com/vc8l01-fauci-admits-on-camera-that-he-lied-about-masks.html}. \\ \vspace{-.2cm}

\noindent {\bf Data:} We constructed our dataset based on original data collected and presented by Biswas et al. \cite{biswas2024political}. They collected over 493k posts shared on Facebook by 5,147 state legislators between 2020 and 2021. To focus on the most misleading content, we refined the ``unreliable'' category \cite{tai2023official}, originally from Media Bias/Fact Check's (MBFC) ratings, by selecting URL domains coded as low or very low factual to create a low-factual reference. Using this reference, we identified low-factual posts. The low-factual post dataset is particularly well-suited for our study, as the shared posts by these politicians are political texts, making them highly relevant to misinformation-related social harms. 

We sampled 500 posts from this dataset. The URLs in the sampled posts linked to text and/or video content. We scraped titles, blurbs, and content in the shared URLs. For videos, we converted them to audio files, and employed Google’s audio-to-text API to generate transcripts. Due to some links being inaccessible or consisting solely of event announcements, we retained 415 available pieces of content. Before hand-coding the content, two human raters independently coded a random set of 30 posts according to our definition of misinformation. The coding achieved a high level of intercoder reliability (Cohen’s Kappa = 0.8). To construct the dataset for rating, we included both link titles and their associated content. When the original content was inaccessible, we combined titles and blurbs. This decision aligns with our definition of misinformation, as well as a recent study on headline-driven sharing \cite{sundar2024sharing}. \\ \vspace{-.2cm}

\noindent {\bf Experiments:} We employed a zero-shot setting to label the credibility of samples. Considering trade-offs between major providers, cost, performance, and open-source versus closed-source frameworks \cite{bail2024can, davidson2024start}, we tested the following models: GPT-4o (OpenAI), Llama 3.1 (Meta; 8B and 70B parameters), Gemma 2 (Google; 9B parameters), and Flan-T5-XL (Google). These models were accessed through their respective APIs: OpenAI’s API for GPT-4o, GROP’s API for Llama 3.1, and Gemma 2 and Flan-T5-XL.

To examine whether summarized information from combined content can help detect low-credibility articles more efficiently, we prompted these five GenAI models to generate summaries. Social media content previews (or user-generated organic posts) of/about articles often include summarized content. Acting as extended headlines or blurbs, these summaries can spread misinformation \cite{sundar2024sharing} and may increasingly be generated by AI. It is essential to evaluate whether GenAI can detect misinformation in such summaries. Unlike using summaries as explanations of fact-checking outcomes \cite{atanasova2020generating}, we used them in content reliability evaluations.

With these tasks, for all models, we employed default temperature parameter of 1.0 and used the following two sets of prompts as our basic prompts to summarize content, rate the credibility of content and provide reasonings, following suggestions on effective prompt instruction \cite{ziems2024can,child2019generating}. Following recent work that uses model-generated explanations to unpack the black box of GenAI models \cite{zeng2024justilm}, we also prompted the models to provide explanations for their ratings. The prompt for summarization reads as follows:

\textit{``You are an assistant that provides a concise and accurate summary of text.''}

The initial task instructions were:

\textit{``Summarize the text accurately and concisely, ensuring the embedded attitude is captured.''}

The system prompt for rating and reasoning reads as follows:

\textit{``You are an assistant that rates the credibility of text from 1 to 5 and provides an explanation for your rating.''}

The initial task instructions were:

\textit{``Rate the level of credibility of the above text on a scale from 1 to 5, where: 
        1 = Very low credibility,
        2 = Low credibility, 
        3 = Medium credibility,
        4 = High credibility,
        5 = Very high credibility. 
        Explain why you gave this rating with the most important reasons.''}
        
We adopted this five-point scale to align with rating schemes used by many fact-checkers, such as Snopes.com, which uses ratings like ``false,'' ``mostly false,'' ``mixed,'' ``almost true,'' and ``true.'' The middle category serves as a borderline classification. We set a threshold of 3 and categorized scores below 3 as low-credibility. We tested thresholds of 2 and 4 for robustness and a threshold of 3 in GPT-4o still provides the best performance. To test prompt sensitivity, we created five variations for the rating and reasoning tasks by modifying terms and requiring the models to consider the underlying meaning of the content, as hidden intentions often carry misleading information \cite{gabriel2021misinfo}. For example, in one variation, we prompted the models to evaluate the intention behind the content.
\vspace{-.2cm}
\section{Results}
Overall, most models followed the majority of prompts to rate content credibility and provide reasoning for their ratings. Except for Flan-T5-XL that consistently refused to provide reasoning for our main prompt, the other models occasionally declined to rate or provide reasoning depending on prompts or content. For example, GPT-4o consistently refused to rate a report about ``deep state'' that contained highly opinionated language in the combined content.  \\ \vspace{-.2cm}

\noindent {\bf Performance Evaluation--Q1:} The results from the hand-coded sample revealed that a majority (65\%) of the low-credibility posts labeled by the domain were genuinely misleading or low-credibility. Performance was evaluated using macro F1 and Matthews Correlation Coefficient (MCC) metrics, averaged across all prompts for each zero-shot model. Results are presented in Table \ref{tab:model_comparison}. GPT-4o demonstrated superior performance with a macro F1 of 0.82 and an MCC of 0.66 when rating combined content. MCC scores are generally lower because they provide a balanced evaluation across all components of the confusion matrix and are less sensitive to class imbalance \cite{chicco2020advantages}. Following GPT-4o, Gemma2 achieved a macro F1 score of 0.80 and an MCC of 0.61. As an open source GenAI model with a relatively smaller parameter size, Gemma2-9b exhibits a comparable performance to the closed source-GPT 4o in rating content accuracy. This suggests that certain open-source GenAI models can be viable alternatives to proprietary models for tasks like content accuracy rating. Regardless of relative better performance of GPT-4o and Gemma2-9b, big variation in performance exists. For example, Flan-T5-XL struggled to assess content reliability with a macro F1 of 0.36 and an MCC of 0.12, which could be attributed to its high sensitivity to prompt design \cite{wei2021finetuned} or its architecture as an encoder-decoder (seq2seq) model designed for structured tasks.

\begin{table}[H]
    \centering
    \resizebox{8.5cm}{!}{  
    \begin{tabular}{l|c|c|c|c|c}
        \hline
        
      & \multicolumn{1}{c|}{\textbf{GPT4o}} & \multicolumn{1}{c|}{\textbf{Llama 3.1-8b}}& \multicolumn{1}{c|}{\textbf{Llama 3.1-70b}} & \multicolumn{1}{c|}{\textbf{Gemma2-9b}} & \multicolumn{1}{c}{\textbf{Flan-T5-XL}}  \\ 
     
        \cline{2-6}
       & content   & content   & content  & content& content   \\ 
        \hline
        F1    & 0.82  & 0.56  & 0.78 & 0.80  &  0.36 \\
        MCC   & 0.66    & 0.36  &  0.59 & 0.61  & 0.12  \\
        \hline
    \end{tabular}
    }
    \caption{Performance Comparison of Different Models on Misinformation Detection Task Using Combined Content} \vspace{-.7cm}
    \label{tab:model_comparison}
\end{table}

\begin{figure}[!h]
    \centering
    \includegraphics[width=1\linewidth]{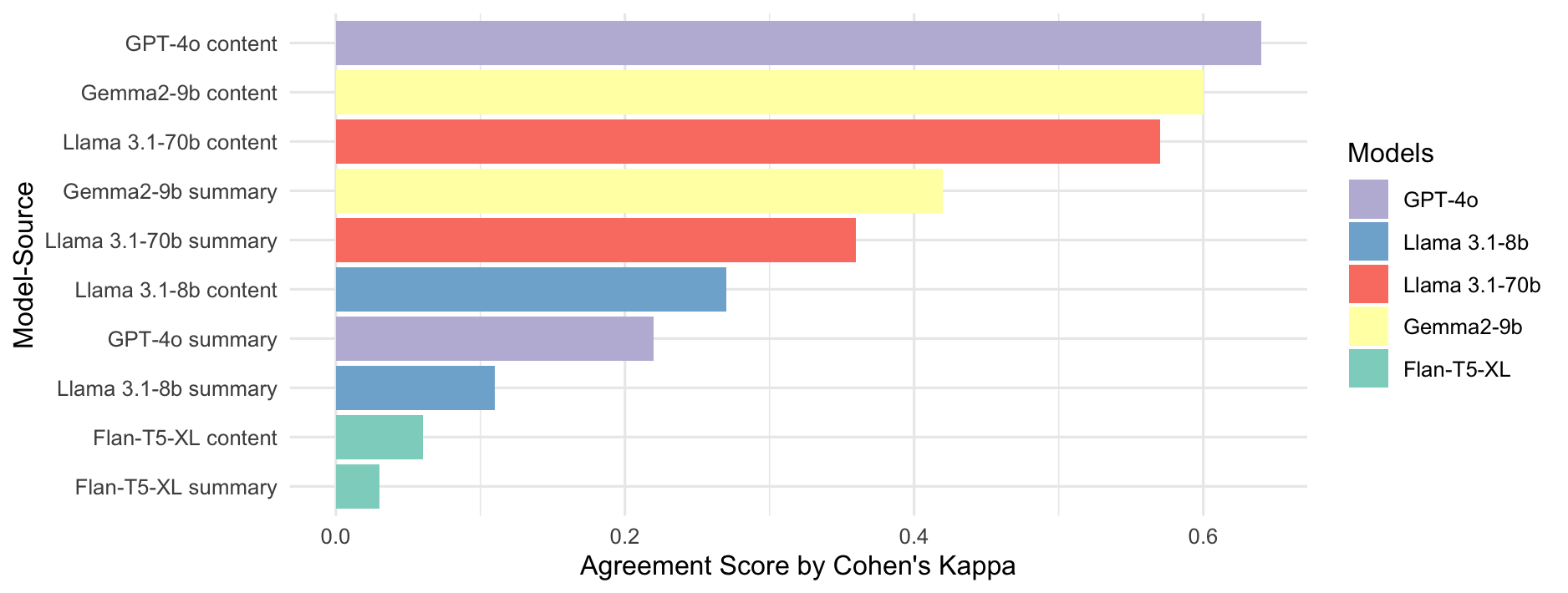}
    \caption{Agreement Score by Model-Source} \vspace{-.25cm}
    \label{fig:kappa}
\end{figure}

\noindent {\bf Matching Human Codes --Q1:} Agreements with human ratings, as shown in Figure \ref{fig:kappa}, were low to moderate across all models. For content-based ratings, the average agreement score was 0.43, with GPT-4o having a moderate agreement at 0.64, followed by Gemma2-9b (0.60) and Llama 3.1-70b (0.57). The remaining models showed much lower agreement, with scores ranging from 0.06 to 0.27. These low agreement scores and the variation among models suggest that the application of GenAI in misinformation detection still has substantial room for improvement. \vspace{-.25cm}

\noindent {\bf Can Summaries Help? --Q2:} On summarized content, Gemma2-9b outperformed others with a macro F1 score of 0.69 and an MCC of 0.49, followed by LLAMA3-70B with a macro F1 score of 0.64 and an MCC of 0.44, as shown in Table \ref{tab:model_comparison_sum}. For the remaining models, F1 scores ranged from 0.34 for Flan-T5-XL to 0.57 for GPT-4o. As shown in Figure \ref{fig:kappa}, the agreement scores with human coders are generally low, with the hightest score of 0.42 by Gemma2-9b and the lowest score of 0.03 by Flan-T5X-XL. The small variation in both performance and agreement scores across models may be attributed to the limited length of the summarized content. 

\begin{table}[H]
    \centering
    \resizebox{8.5cm}{!}{  
    \begin{tabular}{l|c|c|c|c|c}
        \hline
        
      & \multicolumn{1}{c|}{\textbf{GPT4o}} & \multicolumn{1}{c|}{\textbf{Llama 3.1-8b}}& \multicolumn{1}{c|}{\textbf{Llama 3.1-70b}} & \multicolumn{1}{c|}{\textbf{Gemma2-9b}} & \multicolumn{1}{c}{\textbf{Flan-T5-XL}}  \\ 
     
        \cline{2-6}
         & summary  & summary   & summary & summary  & summary  \\ 
        \hline
        F1     & 0.57   & 0.41 & 0.64  & 0.69  & 0.34 \\
        MCC      & 0.39  & 0.21 & 0.44 &0.49   & 0.08  \\
        \hline
    \end{tabular}
    }
    \caption{Performance Comparison of Different Models on Misinformation Detection Task Using Summary} \vspace{-.75cm}
    \label{tab:model_comparison_sum}
\end{table}

To evaluate the quality of summarization, we first calculated Rouge-1 scores \cite{grusky2023rogue} for each model and then examined their relationship with mismatch rates, measured as the proportion of incorrect predictions out of total predictions.

\begin{figure}[htp]
    \centering
    \includegraphics[width=0.9\linewidth]{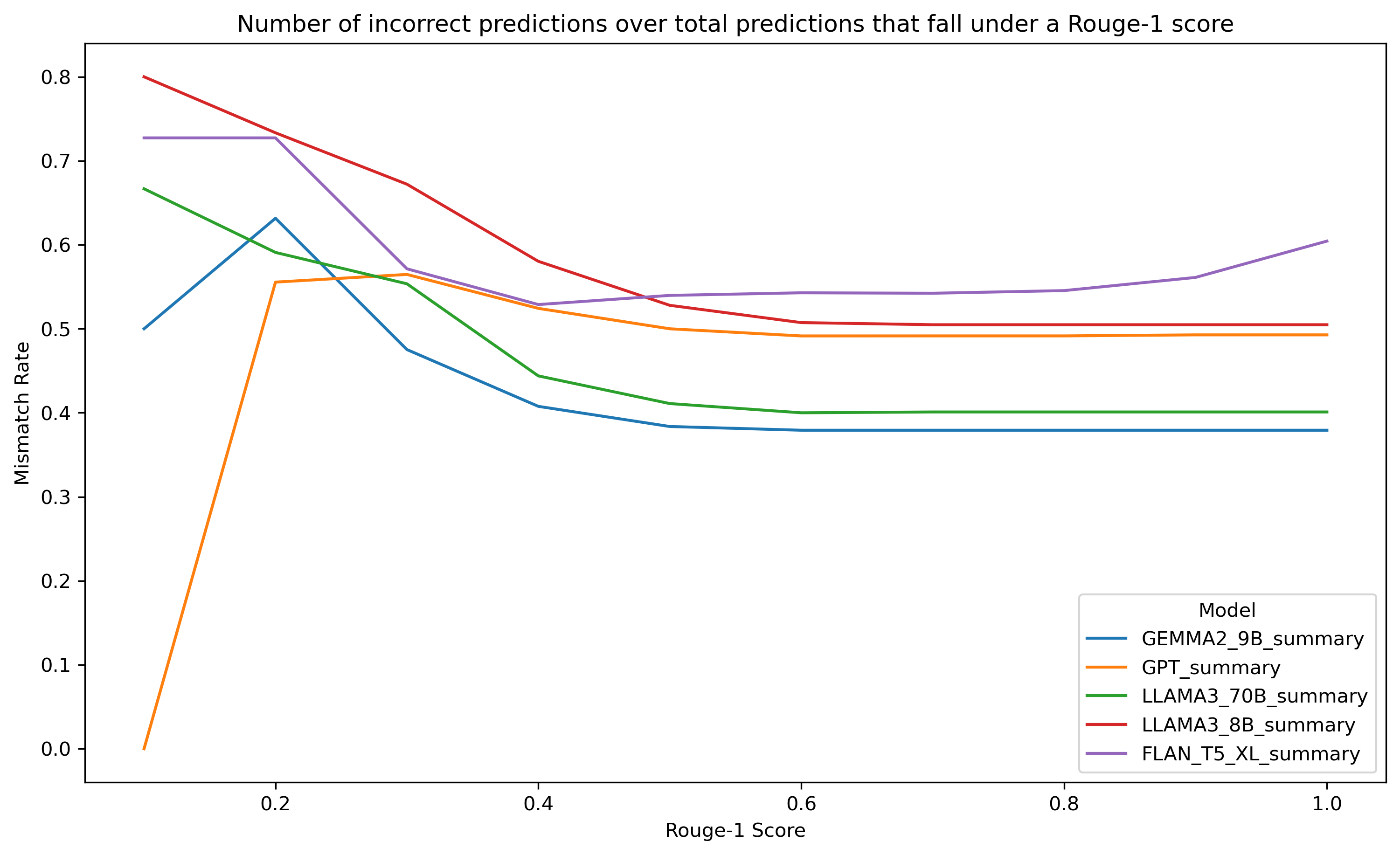}
    \caption{Comparison of the threshold of Rouge-1 scores against the corresponding mismatch rate of LLMs.} \vspace{-.25cm}
    \label{fig:rouge}
\end{figure}

As shown in Figure \ref{fig:rouge}, for most models, an increase in the Rouge-1 threshold generally resulted in a decrease in mismatch rates. Beyond a threshold of 0.5, the mismatch rate stabilized. For example, the mismatch rate for the Llama3.1-70B model decreased steadily from 66.67\% at a threshold of 0.1 to 40.09\% at a threshold of 1.0, indicating consistent improvement with higher thresholds. In contrast, Gemma2-9B exhibited a sharp rise in mismatch rates to 63.15\% at a threshold of 0.2 before decreasing and stabilizing around 37.92\% for thresholds of 0.6 and higher. At these lower thresholds, the model may be misclassifying a larger number of less confident or ambiguous summaries, leading to a higher mismatch rate. When summaries are of higher quality, the performance of models in detecting misinformation using summaries improves. Among all models, Gemma2-9B demonstrated superior summarization capacity, followed by Llama3.1-70B, making them strong candidates for assisting credibility ratings efficiently. 
 \vspace{-.2cm}

\noindent {\bf Assessing GenAI Reasoning Abilities -- Q3:} We analyzed the reasoning provided by models. Except for Flan-T5-XL that refused to provide reasoning for our main prompt, other models generally followed instructions and offered reasoning for their ratings. Given GPT-4o’s superior performance metrics, we focused on its reasoning capacity when evaluating content. We manually reviewed all the explanations provided by GPT-4o and identified three overarching categories and eleven reasoning patterns. The three categories are content quality, language objectivity, and legislative behaviors. These categories are not mutually exclusive and GPT-4o often considers two to three patterns in its reasoning. 

\textbf{Content Quality:} This category emphasizes the quality of the sources, references, and the extent of details provided, as well as the narrative’s balance and linguistic formality.

\textit{Source (Quality of domain source, N = 99)}: This reasoning assesses the credibility/reputation of the domain. For example, content from One America News Network (OANN) is flagged due to a history of ``promoting conspiracy theories and misinformation, especially regarding election integrity.''  according to GPT-4o's reasoning. However, this evaluation sometimes reflects a tendency to infer misinformation based solely on source reputation—even when no explicit connection exists in the text—a form of causal hallucination \cite{diab2024classifying}.

\textit{Details (Details of events and/or evidence for claims, N = 315)}: This reasoning emphasizes whether content includes specific details, such as named individuals, locations, or empirical evidence. GPT-4o also considers whether claims align with known events or have been debunked. This pattern is frequently applied to topics such as election integrity, COVID-19 vaccines, and wearing masks, where substantial fact-checking efforts have been conducted.

\textit{References (Quality of reference sources, N = 176)}: Another reasoning considers whether corroborating details are supported by reliable reference sources. For example, statements from tweets are often flagged as potentially ``misleading or exaggerating.'' 

\textit{Quote (Quality of quote, N = 76)}: The model considers whether quotes come from identified, credible individuals, such as high-ranking officials. For example, GPT-4o often considers direct quotes from public figures as enhancing the credibility of the content. While this aligns with basic principles of credibility assessment, it poses risks of misinformation detection in polarized contexts where political figures often propagate misinformation \cite{tai2023official}.

\textit{Balanced (Balanced narratives and views, N = 59)}: This pattern evaluates whether the content presents balanced narratives or incorporates multiple perspectives, emphasizing objectivity.

\textit{Formality (Language and structure formality, N = 23)}: Language formality assesses structural coherence and linguistic features. Disjointed structures, informal tones, and typographical errors are seen as indicators of diminished reliability. 

\textbf{Language Objectivity:} This category assesses the degree to which the language of the content adheres to objectivity norms.

\textit{Framing (Misleading framing, N = 24)}: Content is flagged as low-credibility when it employs misleading framing. This reasoning could be highly effective, as misleading framing constitutes a significant portion of misinformation \cite{science_misinformation}. However, in our study, this reasoning was applied to only 24 cases, primarily focusing on salient topics of election integrity and COVID-19.

\textit{Biased Language (N = 170)}: This reasoning assesses the objectivity of language, flagging emotive, speculative, or inflammatory language as indicative of low credibility. In contrast, text without sensational elements is considered factual. 

\textit{Opinion (Opinion-based or fact-based, N = 45)}: Content is rated as less credible when it relies on opinion-based statements rather than factual reporting. Opinion pieces authored by politicians are considered to reduce reliability and make it harder to verify facts. This reasoning is problematic in the context of politicians, where opinion-sharing is common.

\textbf{Legislative Behaviors}
This category focuses on content shared by legislators, evaluating whether it reflects genuine legislative activities, such as proposals or procedures, or serves as self-promotion aimed at advancing personal agendas.

\textit{Legislation (Legislative procedure or proposal, N = 14)}: Content credibility is assessed based on its connection to legislative activities, such as proposals or procedures. GPT-4o considers these factors by evaluating whether the described content ``reflect genuine opinions from elected officials'' or align with ongoing policy debates.

\textit{Promotion (Self-promotion and/or personal endorsement, N = 7)}: Self-promotional content is considered less credible because of its agenda-driven nature. While this reasoning is reasonable for assessing credibility in general news reports, it may be less suitable for evaluating public officials, whose communications frequently involve endorsements or agenda-setting.

\begin{figure}[!h]
    \centering
    \includegraphics[width=1\linewidth]{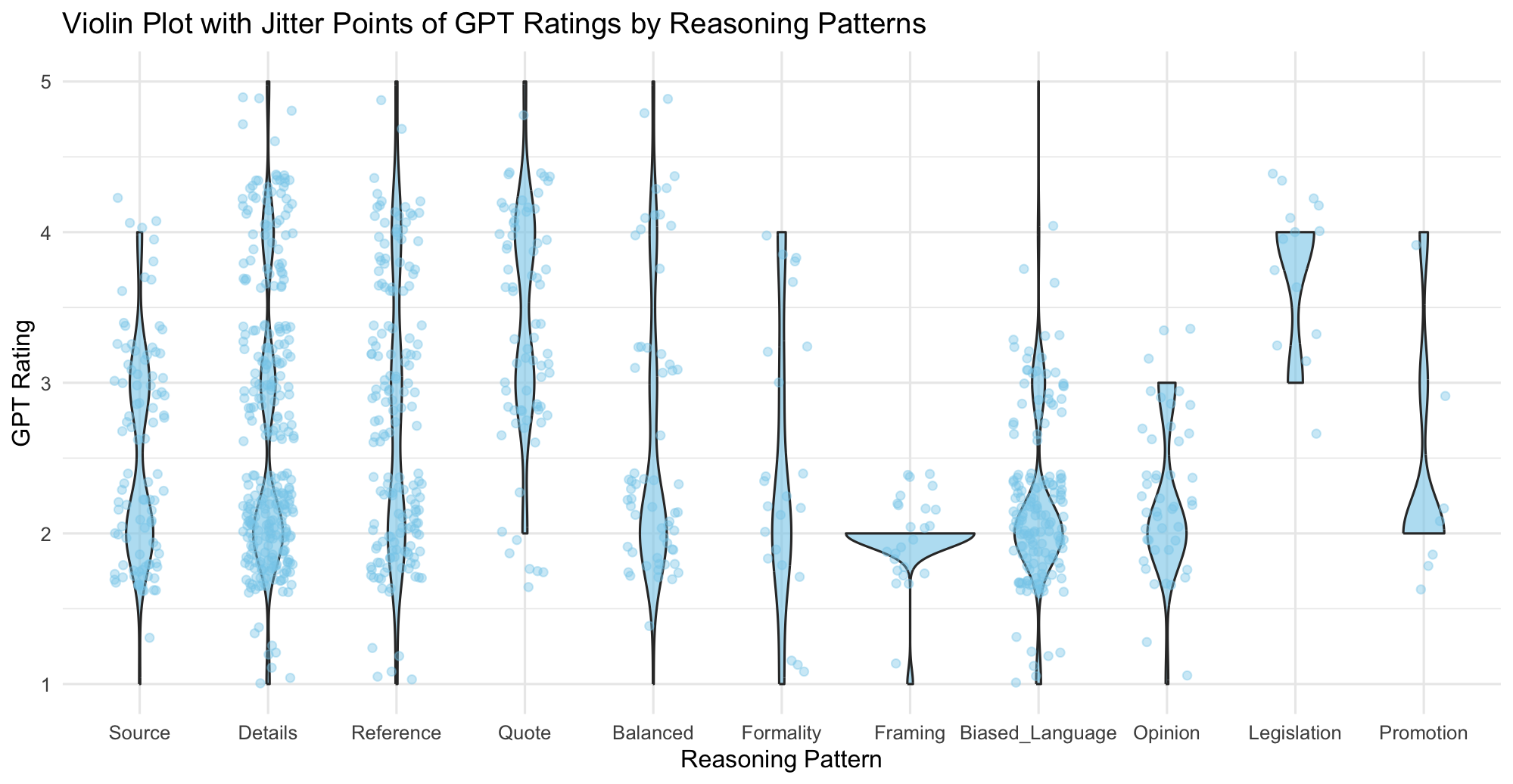}
    \caption{Distribution of GPT Ratings Across Different Reasoning Patterns} \vspace{-.25cm}
    \label{fig:distribution_rates}
\end{figure}

Figure \ref{fig:distribution_rates} gives the distribution of ratings across reasoning patterns. Biased\_Language and Balanced have more variability, with ratings spanning the entire range of 1–5. Source, Details, and Reference patterns exhibit relatively balanced distributions, with ratings clustering around 2–4. Quote pattern shows a concentration of ratings around 4–5, suggesting content evaluated using direct quote is often rated as more credible. In contrast, Framing, Formality and Legislation display a narrower distribution with a density around 2, suggesting more consistent ratings across these patterns and their role in assessing low-credibility content.

\begin{figure}[!htb]
    \centering \vspace{-.35cm}
    \includegraphics[width=1\linewidth,trim=0cm 2cm 0cm 0cm, clip=true]{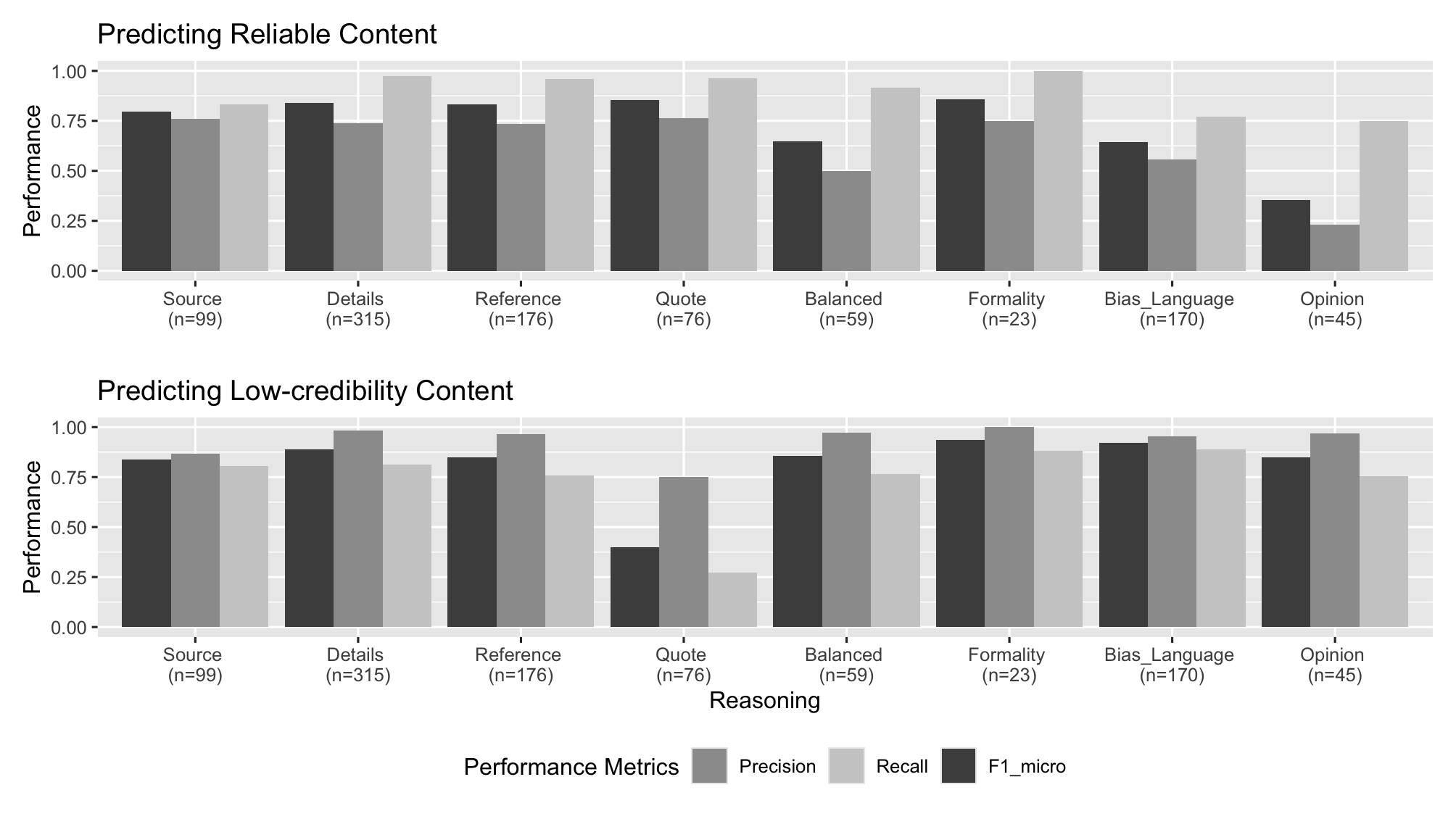}
    \caption{Prediction Performance by Reasoning Patterns} \vspace{-.25cm}
    \label{fig:reason_pattern}
\end{figure}

To better understand the effectiveness of each reasoning pattern, we calculated precision, recall, and F1 scores for both reliable and low-credibility content. We excluded Framing (n = 24), Legislation (n = 14), and Promotion (n = 7) from the analysis on individual reasoning patterns due to the absence of instances in both classes and/or their small sample sizes. In addition to evaluating individual reasoning patterns, we also analyzed the performance of the three broader reasoning groups. The performance metrics for reasoning patterns are presented in Figure \ref{fig:reason_pattern}. 

Source, Details, Reference, and Formality demonstrated consistent performance in predicting both reliable and low-credibility content, achieving moderately high precision, recall, and F1 scores. However, the relatively lower performance in predicting reliable content suggests challenges posed by features that deviate from traditional news report criteria, such as the lack of cross-references, detailed information, and formal language. The presence of these features does not necessarily guarantee accuracy.

Quote showed relatively low effectiveness, particularly in predicting low-credibility content. The model often treats direct quotes from public figures as signals of reliability. However, this approach is problematic, as many public figures and officials have been documented as significant contributors to misinformation \cite{tai2023official}.

In contrast, Biased language, Opinion, and Balanced narratives exhibited better performance in identifying low-credibility content compared to reliable content. The difficulty in labeling reliable content arises from the frequent use of emotive language, one-sided narratives, and opinion-based statements in political texts. 

Among all reasoning patterns, Misleading Framing stands out for its ability to capture nuanced and common features of low-credibility content. All 24 instances where this reasoning was applied were correctly identified as low-credibility. However, the model’s use of this reasoning was limited to specific topics.

\begin{figure}[!htb]
    \centering
 \includegraphics[width=1\linewidth,trim=0cm 2cm 0cm 0cm, clip=true]{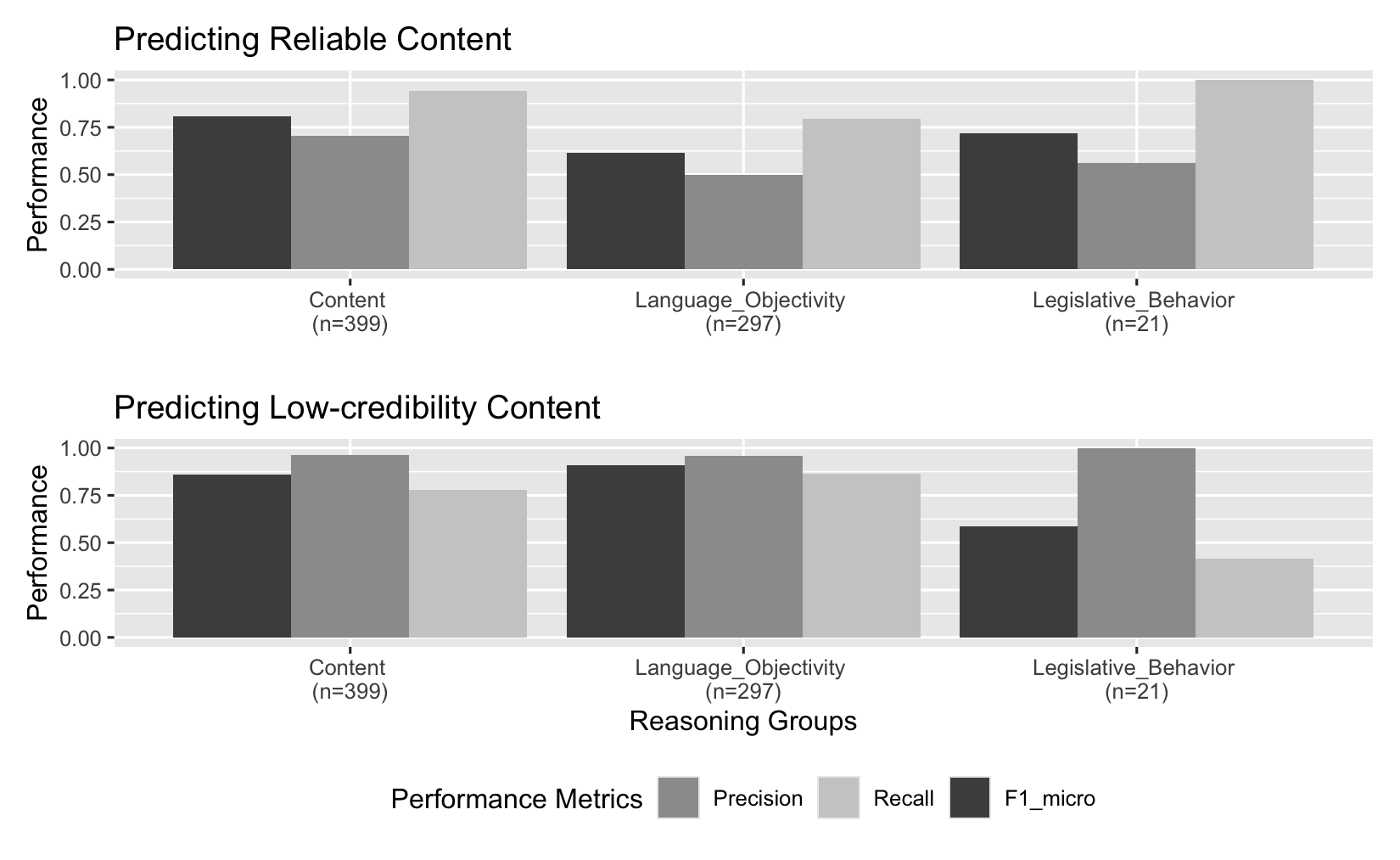} \vspace{-.5cm}
    \caption{Prediction Performance by Reasoning Groups} \vspace{-.25cm}
    \label{fig:reason_groups}
\end{figure}

The performance of three groups of reasoning show a similar pattern in Figure \ref{fig:reason_groups}. The ``hard'' criteria by content quality present a balanced performance in the predictions of both reliable and low-credibility content, with slightly better performance for low-credibility predictions. This suggests that “hard” criteria based on verifiable aspects of content remain reliable. Language objectivity struggled with distinguishing reliable content. Emotive language and subjective framing further complicate accurate predictions of reliable content. The group of legislative behavior is less effective in detecting low-credibility content. Many public officials publicly share misleading or false information in their speeches.

\section{Discussion}

Using a dataset of 500 samples manually coded with high intercoder reliability, we conducted experiments across five GenAI models. These models were evaluated on their ability to rate credibility using both summarized content and combined original content with titles. GenAI models exhibit moderate capacity in rating content credibility, but fall short of matching human accuracy. Among the models tested, GPT-4o demonstrated superior performance when using original content, while Gemma2-9b performed better with summarized content. GenAI models have potential to assist human fact-checkers by scaling the identification of potential misinformation, they cannot be relied upon as standalone tools.

Our examination of GPT-4o’s reasoning revealed three reasoning groups—content quality, language objectivity, and legislative behaviors—comprising eleven specific reasoning patterns. The model predominantly relies on ``hard'' criteria such as the level of detail, source quality, and linguistic features, which are effective in identifying low-credibility content but pose challenges in detecting reliable content. A key concern is whether GenAI models can identify content that appears factual by incorporating cross-references, detailed events, and less biased, less emotive language to manipulate perceptions. This issue is further compounded by the model’s tendency to infer non-existent connections based solely on source reputation and/or the position of mentioned politicians (causal hallucination) \cite{diab2024classifying}. It becomes particularly problematic in contexts where issues have not been widely debunked.

Our study has two important limitations. First, we focused on zero-shot models to assess the capacity of GenAI. While this approach allows us to evaluate the baseline performance of GenAI, it does not leverage alternatives like agentic AI systems with search and retrieval or few-shot learning, which could enhance performance by incorporating external knowledge or task-specific examples. Future studies could explore these alternatives. Second, we categorized reasoning patterns solely for GPT-4o due to its superior performance. Different models may employ distinct reasoning strategies, which are worth further exploration. 

Based on our findings, we recommend that scholars and practitioners should approach GenAI as a complement to, rather than a replacement for, human fact-checkers. A hybrid human-AI teaming approach or human-in-the-loop model is essential for technology-assisted fact-checking. Second, given GenAI’s potential but limited capacity for detecting misinformation, human oversight is critical, particularly for high-stakes tasks.

\vspace{-.2cm}
\begin{acks}
This project is supported by AFOSR, ONR, Minerva, NSF \#2318461,
\#2318460 and \#2148215 awards.
\end{acks}
\vspace{-.2cm}

\bibliographystyle{ACM-Reference-Format}
\bibliography{sample-sigconf}

\end{document}